\documentclass[11pt]{article}
\usepackage{coling2020}
\usepackage{times}
\usepackage{url}
\usepackage{latexsym}

\usepackage{graphicx}
\usepackage{enumitem}
\usepackage{subfig}
\usepackage{tabularx}
\usepackage[T1]{sansmath}
\usepackage{tabulary}

\usepackage{hyperref}
\hypersetup{
    colorlinks=true,
    linkcolor=blue,
    filecolor=magenta,      
    urlcolor=blue,
}

\usepackage{marginnote}
\usepackage{hyperref}

\usepackage[colorinlistoftodos,prependcaption,textsize=tiny]{todonotes}

\colingfinalcopy 

\usepackage{multirow}

\title{A question-answering system for aircraft pilots' documentation}

\author{
  Alexandre ARNOLD\thanks{\hspace{1pt} Authors are in alphabetical order.} \\\and
  {\bf Gérard DUPONT\footnotemark[1]} \\\and
  {\bf Félix FURGER\footnotemark[1]} \\\and
  {\bf Catherine KOBUS\footnotemark[1]} \\\and
  {\bf François LANCELOT\footnotemark[1]} \\ 
  Airbus AI Research \\
  {\tt surname.lastname@airbus.com} \\
}

\date{2020}

\begin{document}
\maketitle
\begin{abstract}
The aerospace industry relies on massive collections of complex and technical documents covering system descriptions, manuals or procedures. This paper presents a question answering (QA) system that would help aircraft pilots access information in this documentation by naturally interacting with the system and asking questions in natural language. After describing each module of the dialog system, we present a multi-task based approach for the QA module which enables performance improvement on a Flight Crew Operating Manual (FCOM) dataset. A method to combine scores from the retriever and the QA modules is also presented.
\end{abstract}


\section{Introduction}
\label{intro}

The aerospace industry relies on large collections of documents covering system descriptions, manuals or procedures. Most of these are subjected to dedicated regulation and/or have to be used in the context of safety of life scenarios such as cockpit procedures for pilots. This documentation, called Flight Crew Operating Manual (FCOM) incorporates aircraft manufacturer guidance on how to use the systems on-board the aircraft for enhanced operational safety, as well as for increased efficiency. Overall it can be seen as several PDF documents which amounts for several thousand pages in total for each aircraft type.

 A user looking for specific information in response to a given situation in this large corpus has to spend a lot of time navigating and searching through the documents. For example, pilots can sometimes have difficulties in finding known items in a constrained time \cite{arnold2019conversational}.

Search technologies are a way to address the structural complexity; however, they come with their own limitations. Most of the time, it is the user's responsibility to define their search needs through specific query syntax and refine the query until the right information is uncovered. This is known as the difficulty of articulating information needs \cite{Liu_2008b} \cite{Wittek2016}. For simple queries that have a ready-made answer in the document, this is not always a difficult problem. However, for the understanding of complex procedures or for troubleshooting system errors, it can lead to multiple queries thus a cumbersome search experience for the user.

The recent advances in natural language understanding and interactive search system have attempted to reduce cognitive overhead. The use of natural language conversation with the system can alleviate the users' need to understand the system's query syntax or document structure. Coupled with high-performing speech-to-text systems, it can even reduce the dependency to physical inputs to free users hand, allowing better multitasking. In this direction, conversational search agents appears to be a promising approach. For a more complete review on data driven dialog systems, please refer to \cite{Serban2015ASO} or \cite{radlinskiTheoreticalFrameworkConversational2017} for a promising framework for conversational search.\\

The release of pretrained language models like BERT~\cite{devlin-etal-2019-BERT} enabled a boost in performance of lots of NLP downstream tasks, in question answering (QA) in particular. It would not have been possible without the emergence of QA datasets too, like SQuAD 2.0~\cite{SQUAD} or CoQA~\cite{reddy-etal-2019-coqa}. Now, best performing systems on those datasets are mainly based on BERT fine-tuning approaches and are nearly approaching human performance.

The main contribution of this work is to present a complete dialog pipeline that allows pilots to access information by naturally interacting with the system. We introduce a multi-task approach for the QA module, which improves the QA performance on a Flight Crew Operating Manual (FCOM) dataset. A method to combine scores from the retriever and the QA modules is also detailed.\\

This paper is organized as follows : in section~\ref{sec:system}, the overall architecture with its different components is detailed, with a focus on the multi-task approach for QA~\ref{sec:qa}. Section~\ref{sec:qa_experiments} describes the dataset used in our experiments and the results obtained. Section~\ref{sec:conclusion} presents a summary of the main findings and future prospects.

While this paper focus on the system side of the study, the reader interested by interactive user study that was executed using this system, is invited to read the dedicated paper \cite{CIRCLE:2020}.

\section{System description}
\label{sec:system}

We have developed a prototype system to address the evaluation objective of determining the relationship between the types of search tasks and the perceived usefulness of search. The system was built around three main components (inspired by the DrQA proposal from \cite{chen2017reading}):

\begin{itemize}
    \item A dialog engine (based on RASA platform~\cite{bocklisch2017rasa}) handling the conversation and identifying user's intents;
    \item A retriever : a search engine (based on Solr \cite{turnbull2016relevant}) where the documents collection is indexed following the BM25F relevance framework~\cite{RoBERTson:2009};
    \item A QA engine, based on a fined-tuned BERT large model \cite{devlin-etal-2019-BERT}. A multi-task setup was used for the fine-tuning: one task is the classical QA task (detecting the span of text) on SQUAD 2.0 dataset~\cite{SQUAD}; the other is a classification task (i.e. whether the answer to the question is contained or not in the document extract). This QA module is described more deeply in the section~\ref{sec:qa}.
\end{itemize}
 
\begin{figure}[!ht]
  \centering
  \includegraphics[width=0.9\columnwidth]{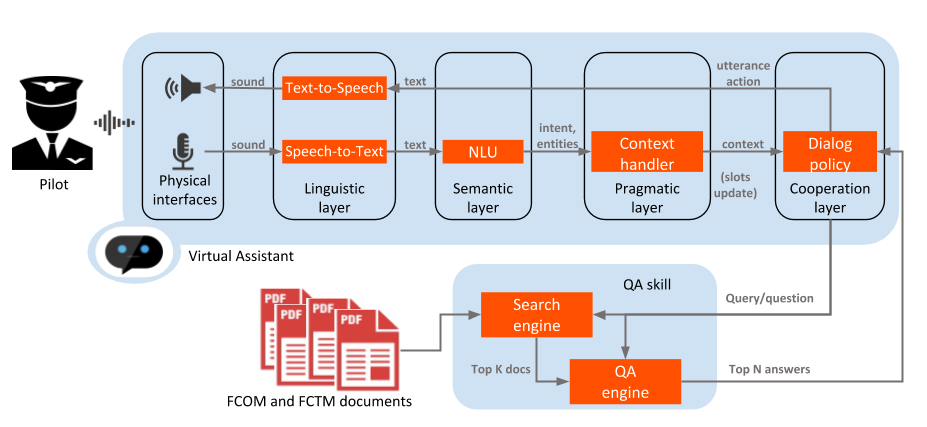}
  \caption{Overview of the prototype architecture.}
  \label{fig:qa_prototype_architecture}
\end{figure}

\begin{figure*}[ht]
  \centering
  \includegraphics[width=0.9 \columnwidth]{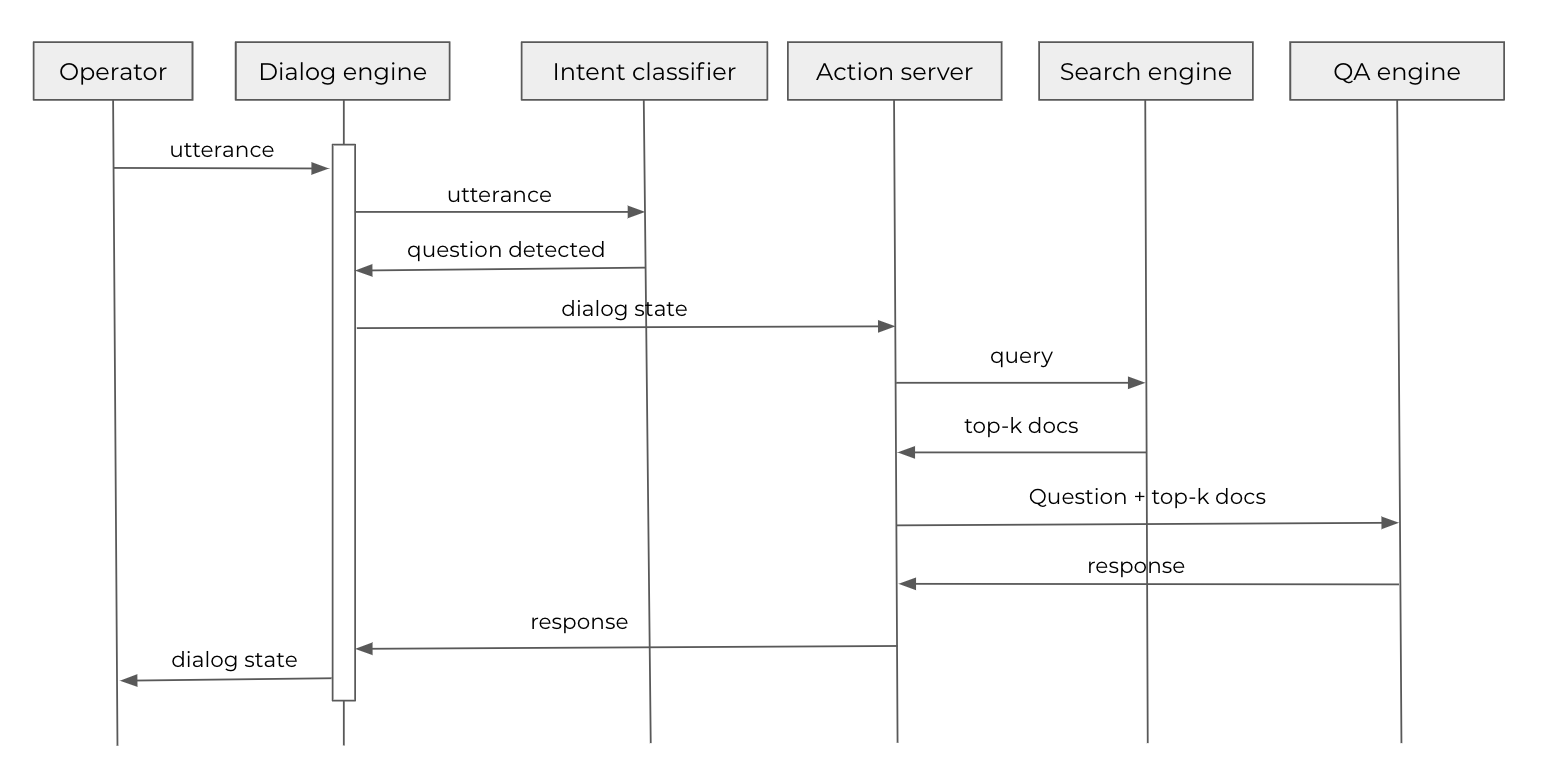}
  \caption{Overall architecture with the dialog model as an orchestrator.}
  \label{fig:overall_architecture}
\end{figure*}

On top of these, additional capabilities to process speech inputs and produce speech outputs are available as an alternative to the traditional textual input. Figures~\ref{fig:qa_prototype_architecture} and~\ref{fig:overall_architecture} offer an overview of the whole architecture.

\begin{figure*}[ht]
  \centering
  \includegraphics[width=0.9\columnwidth]{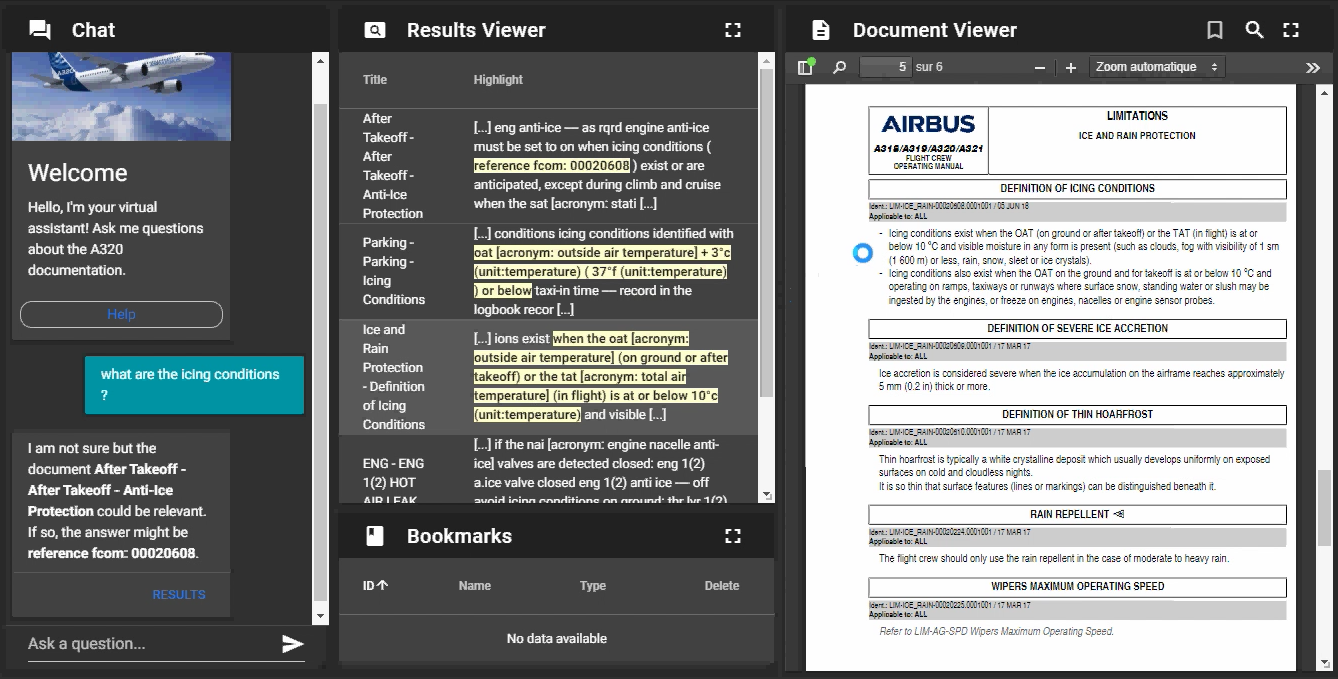}
  \caption{Screenshot of the prototype showing conversation on the left, search result panel in the center and document view on the right.}
  \label{fig:LEA-screenshot}
\end{figure*}

The whole system is made available through a reactive web interface enabling conversation and document exploration (See Figure~\ref{fig:LEA-screenshot}). It was deployed in a cloud environment and made available to users through a tablet.

\subsection{The dialog engine}
We used the open-source Rasa framework for the dialog engine \cite{bocklisch2017rasa}, comprising:

\begin{itemize}
    \item \textbf{Natural language understanding}: recognizing high-level intent/entities from raw user utterances (e.g. "greeting", "positive/negative feedback" or "question")
    \item \textbf{Dialog policy}: predicting the next best action (an utterance or a custom action) based on current dialog state, including last recognized intent
\end{itemize}

Both components above were trained with machine learning pipelines provided in Rasa, based on natural language and story examples: the former maps user utterances to predefined intents/entities, the latter gives typical dialog scenarios to learn \& generalize from (to avoid building manually the whole conversation state machine).

The core "skill" of our dialog engine focuses on recognizing any generic question from the user, mapping it to the "question" intent and predicting the trigger of a custom action (written in Python) which calls the retriever \& QA systems described in next sections to provide an answer. Examples of natural language questions were built by combining open QA dataset questions with in-house examples more related to our pilots' documentation context. We also integrated positive/negative feedback intents to be able to handle user reactions after providing an answer: in case of negative feedback, a custom action is triggered to propose the best answer from the document ranked just below the one currently suggested.

A chitchat "skill" (i.e. another sub-part of our conversational system) was added to the core one, containing more than 50 typical small talk intents \& responses to make the dialog appear more human-like: "greeting", "goodbye", "thanking"... Some chitchat user utterances might be in question form, but are usually learnt not to be confused with the generic "question" intent mentioned above with enough training data.

\subsection{The retriever}
The first component consists of an information retrieval system which follows the now classic architecture of most recent QA systems. It allows to filter the overall document collection 1) to exclude non relevant documents and 2) reduce the considered set of documents to a size that is compatible with the foreseen response time. 

First the document collection has been extracted from its original XML format which includes simultaneously semantic and presentation tagging. The isolation of each individual procedure has been done at this level as well as the extraction of metadata such as unique identifier, applicability scope and classification in the hierarchical ATA chapters\footnote{\url{https://av-info.faa.gov/sdrx/documents/JASC_Code.pdf}}. This allowed to define the minimal granularity of the collection. The text content has then been pre-processed to eliminate inconsistent formatting issues and improve quality of the terms indexed such as resolving abbreviations meanings, adapting the numerical representation of units and flattening table content to ensure headers and legends are correctly indexed.

We compared different indexing scheme and the BM25F from \cite{RoBERTson:2009} allowed to offer the best performances (compared to tf/idf).

\subsection{The QA system}
\label{sec:qa}
\subsubsection{BERT fine-tuning approach}

The QA module in the pipeline is an extractive QA approach, where the model extracts a span of text from a document to answer a natural language question. The QA model is obtained by fine-tuning a BERT language model~\cite{devlin-etal-2019-BERT} for an extractive QA task on the SQuAD 2.0 dataset~\cite{rajpurkar2018know}.

\begin{figure}[t]
  \centering
  \includegraphics[width=\columnwidth]{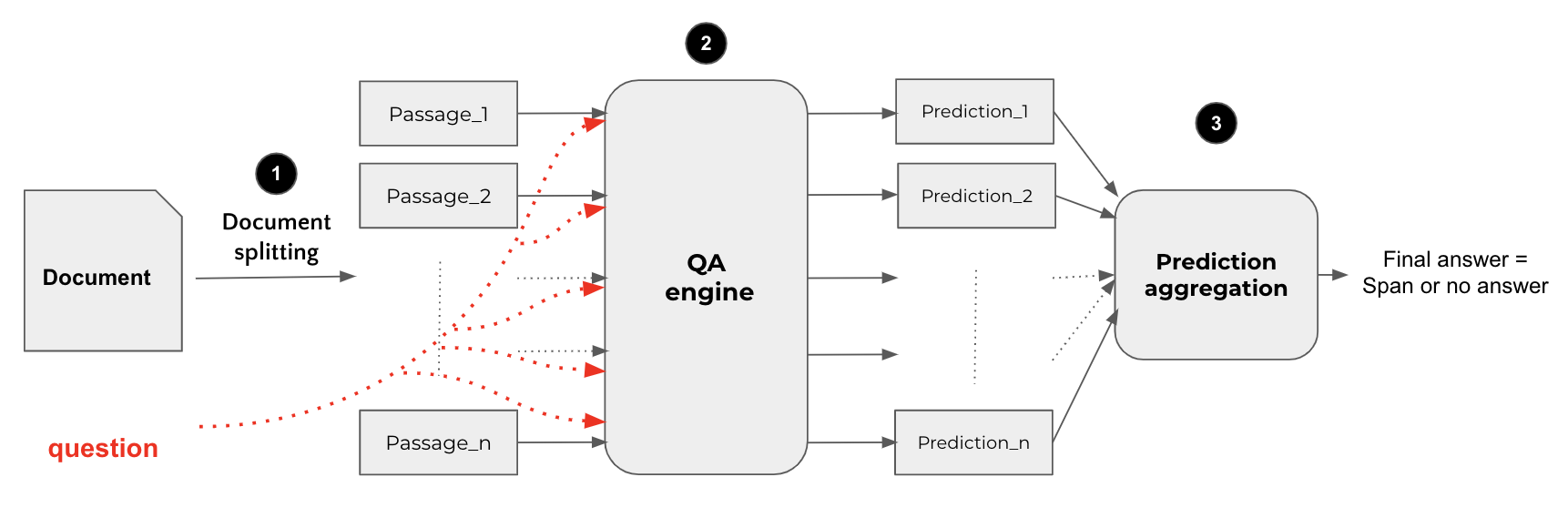}
  \caption{QA engine pipeline}
  \label{fig:qa_pipeline}
\end{figure}

Almost all QA datasets are made of long documents that cannot fit in a standard transformer model. Hence, for a given question and a document to consider, the document is first decomposed into $n$ smaller passages; those passages are provided, with the question to answer, as input to the QA engine; this step results in $n$ predictions, that need to be aggregated to get a final answer for the given question/document pair. The different steps are summarized in figure~\ref{fig:qa_pipeline}.

\begin{figure}
  \centering
  \includegraphics[width=0.5 \columnwidth]{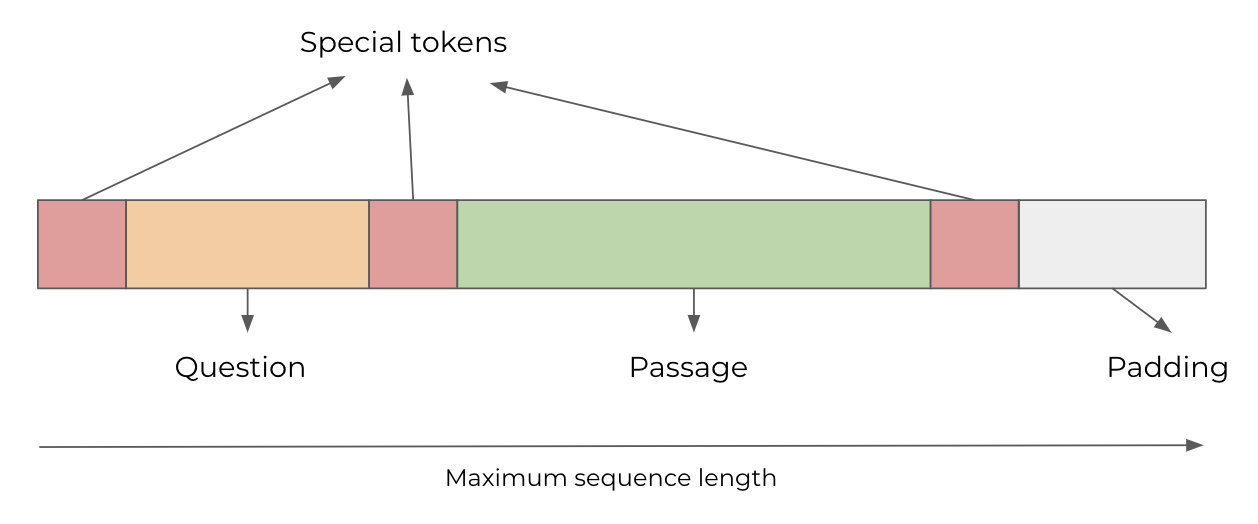}
  \caption{Input to the QA model (with a size equal to the maximum sequence length).}
  \label{fig:QA_input}
\end{figure}

The input~\ref{fig:QA_input} provided to the QA model contains tokens from both the question and the passage, some special tokens and eventually some padding tokens (that enables to reach the model's maximum sequence length if needed).

Formally, we define a training set instance as a triple $(c, s, e)$, where $c$ is a context of a given size ($max\_seq\_len \in \{384, 512\}$ in this study) of wordpiece ids, corresponding to the question, to the passage considered, to some special tokens and eventually some padding. $(s,e) \in [0,max\_seq\_length]^2$ respectively refer to the start and end of the target answer span when the answer span is contained in the passage; they are both equal to $0$ otherwise.

During inference, the predictions for each passage need to aggregated to extract, from the document, the answer (or not) to the question. If the best prediction for each passage is no answer, then the final prediction is no answer. Otherwise, the final prediction is built by picking up the answer span kind prediction with the highest score.

The FARM framework~\footnote{\url{https://github.com/deepset-ai/FARM}} was used to fine-tuned the BERT models on the QA task. More details on the approach can be found in this blog post~\footnote{\url{https://towardsdatascience.com/modern-question-answering-systems-explained-4d0913744097}}.

\subsubsection{Multi-task approach}
\label{sec:MTL}
Multi-Task Learning (MTL) aims at boosting the overall performance of each individual task by
leveraging useful information contained in multiple related tasks. It has shown great success in
Natural Language Processing (NLP). The main idea of MTL is to leverage useful information contained
in multiple related tasks to improve the generalization performance of all the tasks~\cite{survey_MTL_2017}. Multi-task learning has been successfully used in many applications from machine learning, from natural language processing~\cite{ColloBERTMTL_2008} and speech recognition~\cite{Deng_Hinton_2013} to computer vision~\cite{Girshick_2015}, etc.

With the MTL approach and following the formalism proposed in the technical note \cite{DBLP:journals/corr/abs-1901-08634}, a training set instance becomes a $4$-tuple $(c, s, e, t)$ where $c$ is a context of a given size ($max\_seq\_len \in \{384, 512\}$ in this study) of wordpiece ids, corresponding to the question, to the passage considered, to some special tokens and eventually some padding. $(s,e) \in [0,max\_seq\_length]^2$ respectively refer to the start and end of the target answer span. $t$ is the tag associated to the sample with two possible values : $t=\mathbf{SPAN}$ if the answer to the question is in the passage considered, $t=\mathbf{NO\_SPAN}$ otherwise. One can note this is streamlined version compared to the one proposed by \cite{DBLP:journals/corr/abs-1901-08634} which proposed 5 labels for $t$ and showed significant improvements.

Adding this classification head to the network, that tries to predict if it is able to answer a question given the considered passage, should help the overall performance of the QA engine. 

During training, the losses of both tasks, question answering on one hand and classification on the other hand, are summed.
During inference, as for the question answering task, for a question and a given document, all results from all the samples of question/passage have to be aggregated to get at the end a unique classification : "$\mathbf{SPAN}$" if the answer is in the document, "$\mathbf{NO\_SPAN}$". For this step, we went for a basic approach, that consists in taking the classification tag from the passage with the highest score for the classification head.

\section{Experiments}
\label{sec:qa_experiments}

In this study, different BERT models are fine-tuned on the SQuAD 2.0 dataset~\cite{rajpurkar2018know} using the open-source FARM, framework that easily allows to fine-tune BERT model on classical downstream tasks like Question Answering. The different hyper-parameters are synthesized in the table~\ref{tab:QA_hyperparameters}. The code, that enables to fine-tune BERT models, with or without multi-tasking, can be found in the following repository~\footnote{https://github.com/ckobus/FARM}.

\begin{table}
  \begin{center}
    \begin{tabular}{cc}
     & 
     \textbf{BERT large}\\
     \hline
    learning rate & $3 \cdot 10 ^{-5}$\\
    epochs & 2\\
    max. seq. length & $\{384,512\}$\\
    cased & True\\
    \hline
    batch size & $12$\\
    gradient accumulation & $4$\\
    \end{tabular}
    \caption{Values of different hyperparameters used for BERT models fine-tuning.}
    \label{tab:QA_hyperparameters}
   \end{center}
\end{table}

\subsection{Data}
\label{sec:data}
The BERT models, fine-tuned, with or without fine-tuning, on the SQuAD 2.0 dataset~\cite{rajpurkar2018know} are evaluated on both the SQuAD 2.0 development set and on a in-house FCOM dataset built through a crowd-source procedure. Domain experts, knowledgeable about the FCOM or even authors of FCOM procedures, were asked to construct a set of questions/answers, highlighting precisely answers's location in the document. More than $300$ questions were collected but with a large proportion of questions for which the answer is contained in a table or or graphical parts of the document.

\begin{figure}[!ht]
  \centering
  \includegraphics[width=8cm]{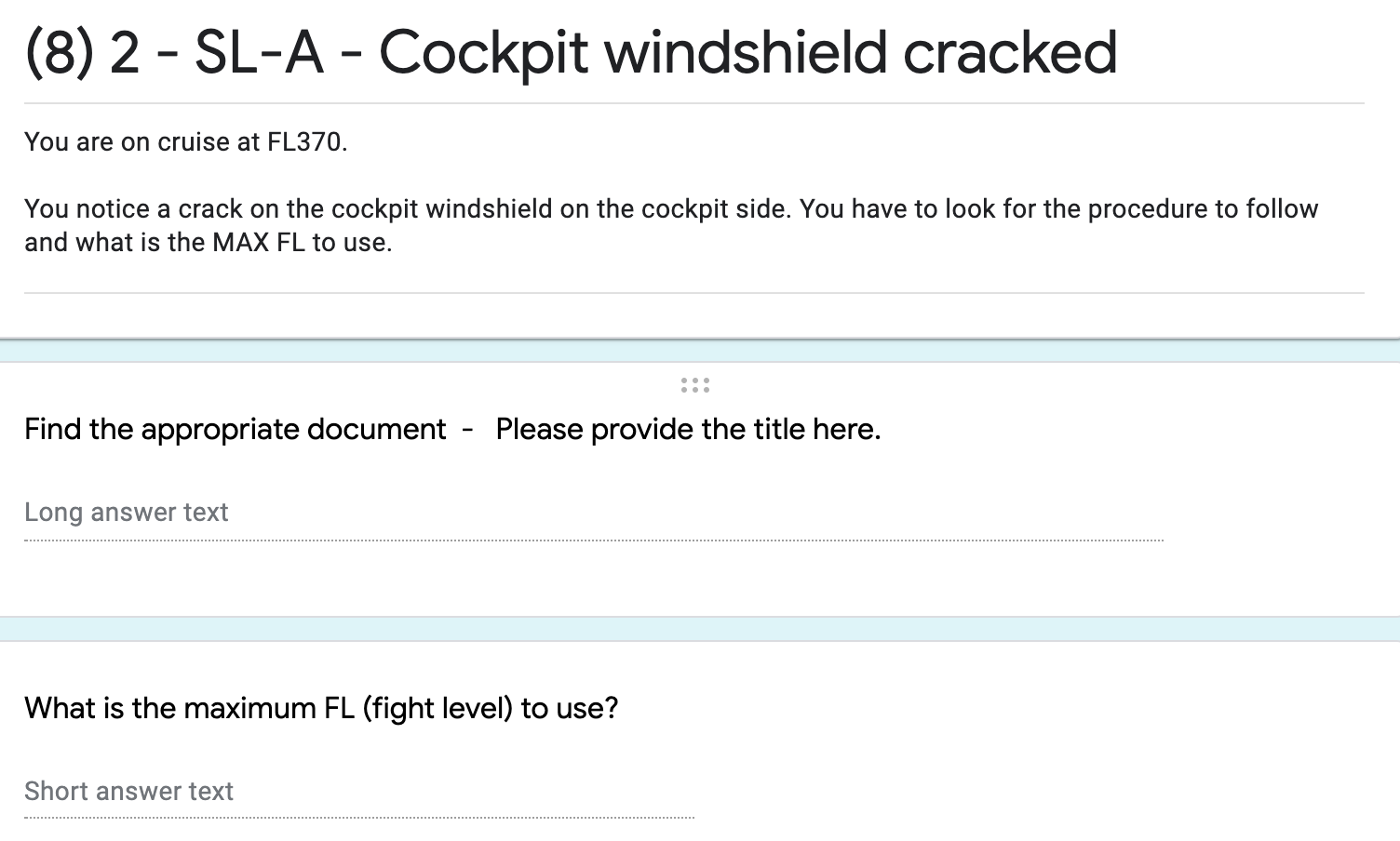}
  \caption{Example of a task.}
  \label{fig:simple_task}
\end{figure}

The dataset was thus filtered so that it only contains questions for which answer is a span of text in the document (see figure \ref{fig:simple_task} for an example). The final version of the FCOM dataset, limited to SQUAD2 type Q/A, contains $114$ questions. The objective was to focus first on this type of mostly factoid questions. This dataset is of course much too limited for any domain training or even fine-tuning. The goal was here to use it for evaluation only to assess applicability of pre-trained literature models to the specificities of the aerospace domain and documents. 

\begin{figure}
  \centering
  \includegraphics[width=1.0 \columnwidth]{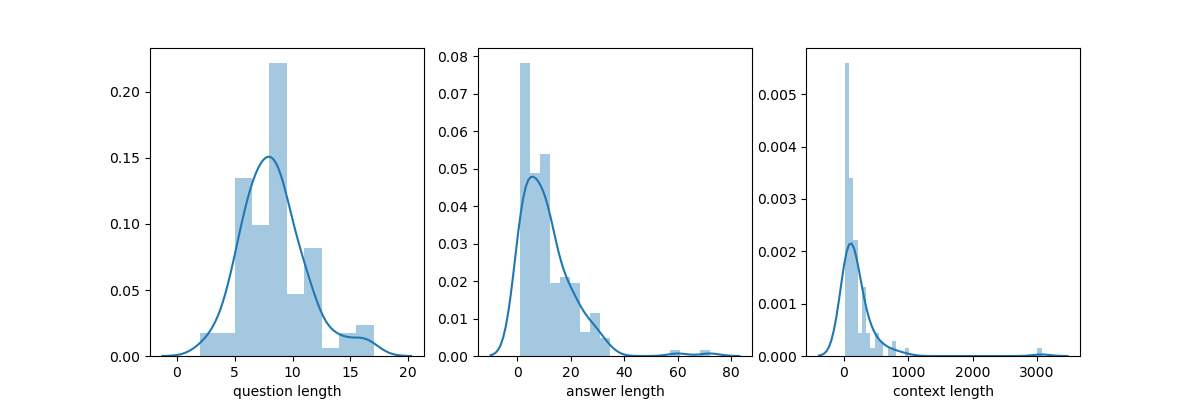}
  \caption{Distribution of the question, answer, and context length.}
  \label{fig:fcom_dataset_distribution}
\end{figure}

The distribution of the questions, answers, and context length of the FCOM dataset can be seen in figure ~\ref{fig:fcom_dataset_distribution}. The average length of questions (in terms of word tokens) is less than 9, while the average length of answers is 11. The average length of context is 200 but some context sentences can be long (3072 words for example).

\subsection{QA performance}
The results obtained with the different BERT large models (with or without whole word masking) on the 2 SQuAD datasets and on the FCOM internal dataset are reported in terms of Exact Match (EM) and F1 scores, in table~\ref{tab:QA_results}. The results for the DrQA model (only the document reader part - see \cite{chen2017reading}) have been reproduced and are presented for comparison.

Multiple conclusions can be drawn from those results :
\begin{itemize}
    \item While the multi-task approach brings limited improvements on the SQuAD 2.0 dataset, the improvement is quite significant in terms of both $EM$ and $F1$ scores on the FCOM dataset, for which the baseline performance is also quite low compared to the SQuAD dataset; this result is expected because the FCOM dataset is made of complex aeronautical documents, with a technical vocabulary and specific phraseology, that is obviously not found in the SQuAD 2.0 dataset, on which the BERT model is fine-tuned; the multi-task approach enables to get in average $18\%$ and $15\%$ respectively in terms of $EM$ and $F1$ on the FCOM dataset;
    \item The maximum sequence length hyperparameter has a limited impact on the SQuAD dataset whereas it has a significant one on the FCOM dataset. The paragraphs in the FCOM dataset are larger (in average $200$ words and some paragraphs can contain more than $3k$ words) than the ones in the SQuAD datasets. Having a larger maximum sequence length enables the QA model to capture a larger context, which helps it in getting a better aggregated answer in the end;
    \item The best performance are obtained with the \textit{whole word masking} version of BERT large model, which confirms the trend in the NLP community~\cite{DBLP:journals/corr/abs-1906-08101}. Hence, this wmm version of the model is used for the experiments about retiever and QA scores combination in the following section~\ref{sec:combination_experiments}.
\end{itemize}


\begin{table}
  \begin{center}
    \begin{tabular}{c|c|c| c c | c c} 
      \textbf{Model} & \textbf{max\_seq\_length} & \textbf{Multi-task} &\multicolumn{2}{c}{\textbf{SQUAD 2.0}} & \multicolumn{2}{c}{\textbf{FCOM}}\\
       & & & EM & F1 & EM & F1\\ 
      \hline
      \hline
      DrQA & NA & No & $34.25$ & $39.21$ & $28.95$ & $48.57$\\
      \hline
      \multirow{4}{*}{BERT large} & \multirow{2}{*}{$384$} & No & $77.54$ & $80.55$ & $21.92$ & $30.08$\\
      \cline{3-7}
       &  & Yes & $76.52$ & $79.86$ & $24.56$ & $34.40$\\
       \cline{2-7}
       & \multirow{2}{*}{$512$} & No & $76.17$ & $79.27$& $23.68$ & $33.62$\\
       \cline{3-7}
       & & Yes & $76.70$ & $79.93$ & $29.82$ & $39.38$\\
       \hline
       \multirow{2}{*}{BERT large (whole word masking)} & \multirow{2}{*}{$512$} & No & $81.60$ & $84.28$ & $29.82$ & $40.77$\\
       \cline{3-7}
        & & Yes & \textbf{82.10} & \textbf{84.98} & \textbf{35.08} & \textbf{46.14}\\
    \end{tabular}
    \caption{QA systems evaluation on SQUAD, SQUAD 2.0 (development set) and on FCOM datasets, without and with multi-tasking. DrQA relies on RNN so no specific limit on sequence length and it was designed for the original SQUAD (with no rejection) on which it obtains $69.08$ and $78.47$ in EM and F1.
    }
    \label{tab:QA_results}
  \end{center}
\end{table}

\subsection{Retriever and QA scores combination}
\label{sec:combination_experiments}

In order to improve document ranking as given by the retriever alone (Solr), we propose to also leverage the confidence score returned by the QA system (BERT) for the document's best answer span. The intuition is that a high confidence in best answer span should partly reflect the document's relevance with regard to the question asked, thus an indication of a higher ranking. We first introduce two simple baselines to combine retriever and QA scores into one, which will serve to re-rank documents by sorting in ascending order:

\begin{itemize}
    \item \textbf{\textit{Simple combination 1}}: retriever and QA scores are multiplied (multiplication is preferred over addition here, since both scores have different scales and orders of magnitude)
    \item \textbf{\textit{Simple combination 2}}: retriever and QA z-scores, computed from absolute scores, are added (z-scores give a normalized relative scoring taking into account other results in the set, thus might be more meaningful than raw scores to know how good a document or answer is compared to the others)
\end{itemize}

Additionally we introduce a more advanced \textbf{\textit{XGBoost combination}}, predicting the re-ranking score using XGBoost, an efficient machine learning framework based on gradient boosting. This has the ability to learn more complex combinations than the linear interpolation proposed in~\cite{yang2019end}. Also we propose to take as input features not only the raw scores of the retriever and QA system, but their respective z-scores as well (as mentioned in the \textit{Simple combination 2}). The XGBoost model is trained with 100 rounds on 80\% of the FCOM dataset and tested on the remaining 20\% (this in-house dataset includes both the expected document and answer span in the ground truth). We compare the different ranking systems with the nDCG@10 metric (normalized Discounted Cumulative Gain accumulated at rank position 10), which is common for document ranking evaluations.

Results are shown in table~\ref{tab:Doc_ranking}; they were obtained with the multi-task QA model, fine-tuned from the BERT large model obtained with whole word masking and with a maximum sequence size of $512$ (which gave the best results overall
~\ref{tab:QA_results}). For the XGBoost part, more rounds of training (e.g. 300) or additional input features (e.g. original document rank, question length...) did not improve performance further - on the latter, it was even detrimental because of over-fitting effects.

\begin{table}
  \begin{center}
    \begin{tabular}{l|c}
        \textbf{Ranker}      & \textbf{\begin{tabular}[c]{@{}c@{}}Mean nDCG@10\\ (on test set)\end{tabular}} \\ \hline
        Retriever alone      & 0.86                                                                          \\
        QA system alone      & 0.68                                                                          \\
        Simple combination 1 & 0.75                                                                          \\
        Simple combination 2 & 0.83                                                                          \\
        XGBoost combination  & \textbf{0.97}                                                                
    \end{tabular}
    \caption{Evaluation of document ranking methods on FCOM dataset.}
    \label{tab:Doc_ranking}
  \end{center}
\end{table}

\begin{figure}[t]
  \centering
  \includegraphics[width=\columnwidth]{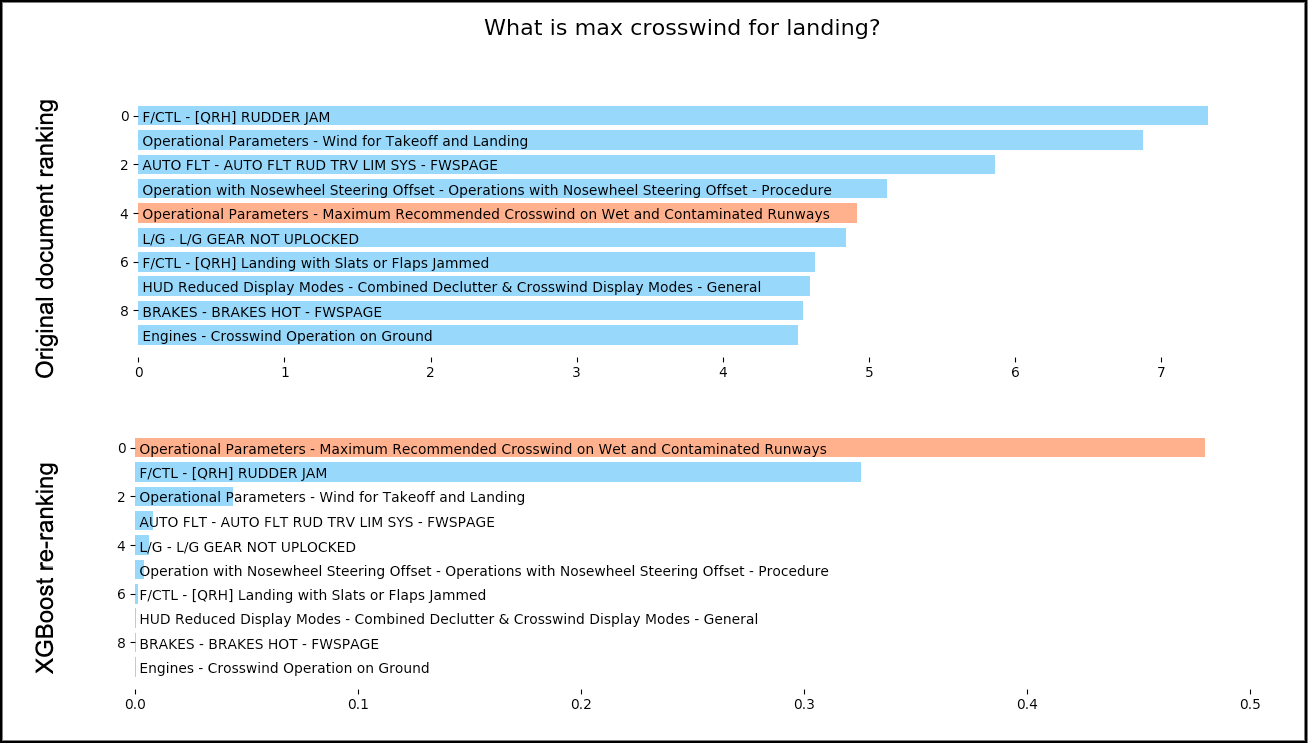}
  \caption{Example of XGBoost document re-ranking benefit}
  \label{fig:reranking}
\end{figure}

An example of the XGBoost document re-ranking is shown in figure~\ref{fig:reranking} for the question "What is max crosswind for landing?". For each scoring configuration (pure document ranking vs XGBoost reranking), the top 10 answers are listed with horizontal bars representing their scores. The right document containing the answer, highlighted in red, is correctly ranked 1st with XGBoost, as opposed to 5th with the original retriever ranking on this example.

Many possible re-ranking could be explored to improve the performance of this component . However these might necessitate larger training corpora. Given the small size of the FCOM dataset (only few hundreds Q/A pairs), it was chosen not to focus too much on this single component of the system.

\subsection{System overall performance}

An interactive user experimentation was proposed to confirm the positive perceptions of the offline performances measured. Our evaluation objective was to \emph{determine the relationship between the search tasks in the typical flight operation scenarios and the perceived usefulness of the system for task completion} (see relevant literature on information-seeking strategies and perceived usefulness of information resources \cite{Earle:2015,Freund2013,Hertzum2019,Li2008a}). 

The experiment was conducted in the environment of a flight simulator (ENAC BIGONE A320/A330 cockpit simulator) within the ACHIL platform. The setting was intended to create an environment that can elicit the information needs of participants, as suggested in simulated work task situations~\cite{borlund2016study}. The subjects were given access to a tablet - similar to the ones used by the pilot in flight - to access the Flight Crew Operating Manual (FCOM) through one of the two systems: our system, called Smart Librarian (SL) and electronic flight bag (EFB), which is the system used currently by pilots and is basically pdf viewer with a keyword search functionality.

Several search tasks with different complexity were designed for the user experiment. Specifically, the easy task involves fact-finding (see figure \ref{fig:simple_task}) while the hard task requires a higher level of understanding of the problems and/or some cognitive reasoning for answering the questions. In easy search tasks, the problem description contains relevant words that can be used to craft the "best question" pointing to a unique procedure (or document unit) that contains the solution. By contrast, in hard search tasks, the problem description does not contain any words matching the "best question" and the subject will need to rephrase the problem. Moreover, the user needs to explore at least two document units to find the answer. 

One of the conclusion is that our system (SL) was more helpful than the current system (EFB) for the more complex search tasks. A complete description of this user study including data analysis and conclusion about the experiment are detailed in a dedicated paper \cite{CIRCLE:2020}.

\section{Conclusion and Perspectives}
\label{sec:conclusion}

In this paper, we proposed a multi-task approach for Question Answering, which enables to improve QA engine's performance especially in a technical domain like the aeronautic one. We also proposed an innovative way to combine scores from both the retriever and the QA module, which enables a combination of the two modules.

There are a number of obvious improvements we might consider; a first obvious one is to get more in-domain data, that will allow us to further fine-tune the QA engine on in-domain data. 

With respect to the multi-task approach, the way the predictions are aggregated to get the final classification tag is a baseline and can be improved.

An other point of improvement is related to tables handling; FCOM documentations are technical and contain a lot of tables. QA engines should be able to handle those tables and extract answer from them if needed. There are already promising works around that topic~\cite{yin2020taBERT}
~\cite{herzig2020tapas} and around the recent Natural Question dataset~\cite{NQ}.

\bibliographystyle{coling}
\bibliography{arxiv2020}

\begin{thebibliography}{}

\bibitem[\protect\citename{Alberti \bgroup et al.\egroup
  }2019]{DBLP:journals/corr/abs-1901-08634}
Chris Alberti, Kenton Lee, and Michael Collins.
\newblock 2019.
\newblock A {BERT} baseline for the natural questions.
\newblock {\em CoRR}, abs/1901.08634.

\bibitem[\protect\citename{Arnold \bgroup et al.\egroup
  }2019]{arnold2019conversational}
Alexandre Arnold, G{\'e}rard Dupont, Catherine Kobus, and Fran{\c{c}}ois
  Lancelot.
\newblock 2019.
\newblock Conversational agent for aerospace question answering: A position
  paper.
\newblock In {\em Proceedings of the 1st Workshop on Conversational Interaction
  Systems (WCIS at SIGIR). Paris}.

\bibitem[\protect\citename{Bocklisch \bgroup et al.\egroup
  }2017]{bocklisch2017rasa}
Tom Bocklisch, Joey Faulkner, Nick Pawlowski, and Alan Nichol.
\newblock 2017.
\newblock Rasa: Open source language understanding and dialogue management.

\bibitem[\protect\citename{Chen \bgroup et al.\egroup }2017]{chen2017reading}
Danqi Chen, Adam Fisch, Jason Weston, and Antoine Bordes.
\newblock 2017.
\newblock Reading wikipedia to answer open-domain questions.

\bibitem[\protect\citename{Cui \bgroup et al.\egroup
  }2019]{DBLP:journals/corr/abs-1906-08101}
Yiming Cui, Wanxiang Che, Ting Liu, Bing Qin, Ziqing Yang, Shijin Wang, and
  Guoping Hu.
\newblock 2019.
\newblock Pre-training with whole word masking for chinese {BERT}.
\newblock {\em CoRR}, abs/1906.08101.

\bibitem[\protect\citename{Deng~L.}2013]{Deng_Hinton_2013}
Kingsbury~B. Deng~L., Hinton G.~E.
\newblock 2013.
\newblock New types of deep neural network learning for speech recognition and
  related applications: An overview.
\newblock In {\em 2013 IEEE International Conference on Acoustics, Speech and
  Signal Processing}, pages 8599–--8603.

\bibitem[\protect\citename{Devlin \bgroup et al.\egroup
  }2019]{devlin-etal-2019-BERT}
Jacob Devlin, Ming-Wei Chang, Kenton Lee, and Kristina Toutanova.
\newblock 2019.
\newblock {BERT}: Pre-training of deep bidirectional transformers for language
  understanding.
\newblock In {\em Proceedings of the NAACL-HLT 2019}, pages 4171--4186,
  Minneapolis, Minnesota. Association for Computational Linguistics.

\bibitem[\protect\citename{Freund}2013]{Freund2013}
Luanne Freund.
\newblock 2013.
\newblock {A cross-domain analysis of task and genre effects on perceptions of
  usefulness}.
\newblock {\em Inf. Process. Manage.}, 49(5):1108--1121.

\bibitem[\protect\citename{Girshick}2015]{Girshick_2015}
R.~Girshick.
\newblock 2015.
\newblock Fast r-cnn.
\newblock In {\em In Proceedings of the IEEE International Conference on
  Computer Vision}, page 1440–1448.

\bibitem[\protect\citename{H \bgroup et al.\egroup }2015]{Earle:2015}
Earle~Ralph H, Rosso~Mark A, and Alexander~Kathryn E.
\newblock 2015.
\newblock User preferences of software documentation genres.
\newblock In {\em Proceedings of the Annual International Conference on the
  Design of Communication}, SIGDOC '15, pages 46:1--46:10, New York. ACM.

\bibitem[\protect\citename{Hertzum and Simonsen}2019]{Hertzum2019}
Morten Hertzum and Jesper Simonsen.
\newblock 2019.
\newblock How is professionals' information seeking shaped by workplace
  procedures? a study of healthcare clinicians.
\newblock {\em Inf. Process. Manage.}, 56(3):624--636.

\bibitem[\protect\citename{Herzig \bgroup et al.\egroup }2020]{herzig2020tapas}
Jonathan Herzig, Pawe{\l}~Krzysztof Nowak, Thomas M{\"u}ller, Francesco
  Piccinno, and Julian~Martin Eisenschlos.
\newblock 2020.
\newblock Tapas: Weakly supervised table parsing via pre-training.
\newblock {\em arXiv preprint arXiv:2004.02349}.

\bibitem[\protect\citename{Kwiatkowski \bgroup et al.\egroup }2019]{NQ}
Tom Kwiatkowski, Jennimaria Palomaki, Olivia Redfield, Michael Collins, Ankur
  Parikh, Chris Alberti, Danielle Epstein, Illia Polosukhin, Matthew Kelcey,
  Jacob Devlin, Kenton Lee, Kristina~N. Toutanova, Llion Jones, Ming-Wei Chang,
  Andrew Dai, Jakob Uszkoreit, Quoc Le, and Slav Petrov.
\newblock 2019.
\newblock Natural questions: a benchmark for question answering research.
\newblock {\em Transactions of the Association of Computational Linguistics}.

\bibitem[\protect\citename{Liu \bgroup et al.\egroup }2020]{CIRCLE:2020}
Ying-Hsang Liu, Alexandre Arnold, Gérard Dupont, Catherine Kobus, and
  François Lancelot.
\newblock 2020.
\newblock Evaluation of conversational agents for aerospace domain.
\newblock In {\em To appear in Joint Conference of the Information Retrieval
  Communities in Europe, CIRCLE 2020}.

\bibitem[\protect\citename{Pengcheng \bgroup et al.\egroup
  }2020]{yin2020taBERT}
Yin Pengcheng, Neubig Graham, Yih Wen-tau, and Riedel Sebastian.
\newblock 2020.
\newblock Tabert: Pretraining for joint understanding of textual and tabular
  data.
\newblock {\em arXiv preprint arXiv:2005.08314}.

\bibitem[\protect\citename{Pia and David}2016]{borlund2016study}
Borlund Pia and Bawden David.
\newblock 2016.
\newblock A study of the use of simulated work task situations in interactive
  information retrieval evaluations: A meta-evaluation.
\newblock {\em Journal of Documentation}.

\bibitem[\protect\citename{R. and J.}2008]{ColloBERTMTL_2008}
Collobert R. and Weston J.
\newblock 2008.
\newblock A unified architecture for natural language processing.
\newblock In {\em Proceedings of the 25th International Conference on Machine
  Learning - ICML ’08}, pages 160--167, New York, NY. ACM.

\bibitem[\protect\citename{Radlinski and
  Craswell}2017]{radlinskiTheoreticalFrameworkConversational2017}
Filip Radlinski and Nick Craswell.
\newblock 2017.
\newblock A {{Theoretical Framework}} for {{Conversational Search}}.
\newblock In {\em Proceedings of the 2017 {{Conference}} on {{Conference Human
  Information Interaction}} and {{Retrieval}} - {{CHIIR}} '17}, Oslo, Norway.
  {ACM Press}.

\bibitem[\protect\citename{Rajpurkar \bgroup et al.\egroup }2018a]{SQUAD}
Pranav Rajpurkar, Robin Jia, and Percy Liang.
\newblock 2018a.
\newblock Know what you don{'}t know: Unanswerable questions for {SQ}u{AD}.
\newblock In {\em Proceedings of the Annual Meeting of the ACL}, pages
  784--789, Melbourne, Australia. Association for Computational Linguistics.

\bibitem[\protect\citename{Rajpurkar \bgroup et al.\egroup
  }2018b]{rajpurkar2018know}
Pranav Rajpurkar, Robin Jia, and Percy Liang.
\newblock 2018b.
\newblock Know what you don't know: Unanswerable questions for squad.

\bibitem[\protect\citename{Reddy \bgroup et al.\egroup
  }2019]{reddy-etal-2019-coqa}
Siva Reddy, Danqi Chen, and Christopher~D. Manning.
\newblock 2019.
\newblock {C}o{QA}: A conversational question answering challenge.
\newblock {\em Transactions of the Association for Computational Linguistics},
  7:249--266, March.

\bibitem[\protect\citename{Robertson and Zaragoza}2009]{RoBERTson:2009}
Stephen Robertson and Hugo Zaragoza.
\newblock 2009.
\newblock The probabilistic relevance framework: {BM25} and beyond.
\newblock {\em Found. Trends Inf. Retr.}, 3(4):333--389.

\bibitem[\protect\citename{Serban \bgroup et al.\egroup }2015]{Serban2015ASO}
Iulian Serban, Ryan~Joseph Lowe, Peter Henderson, Laurent Charlin, and Joelle
  Pineau.
\newblock 2015.
\newblock A survey of available corpora for building data-driven dialogue
  systems.
\newblock {\em CoRR}, abs/1512.05742.

\bibitem[\protect\citename{Turnbull and Berryman}2016]{turnbull2016relevant}
Doug Turnbull and John Berryman.
\newblock 2016.
\newblock {\em Relevant Search: With Applications for Solr and Elasticsearch}.
\newblock Manning Publications, Shelter Island, NY.

\bibitem[\protect\citename{Wittek \bgroup et al.\egroup }2016]{Wittek2016}
Peter Wittek, Ying-Hsang Liu, S{\'{a}}ndor Dar{\'{a}}nyi, Tom Gedeon, and
  Ik~Soo Lim.
\newblock 2016.
\newblock {Risk and ambiguity in information seeking: Eye gaze patterns reveal
  contextual behavior in dealing with uncertainty}.
\newblock {\em Front. Psychol.}, 7:1790.

\bibitem[\protect\citename{Yang \bgroup et al.\egroup }2019]{yang2019end}
Wei Yang, Yuqing Xie, Aileen Lin, Xingyu Li, Luchen Tan, Kun Xiong, Ming Li,
  and Jimmy Lin.
\newblock 2019.
\newblock End-to-end open-domain question answering with bertserini.
\newblock {\em arXiv preprint arXiv:1902.01718}.

\bibitem[\protect\citename{Ying-Hsang and J}2008]{Liu_2008b}
Liu Ying-Hsang and Belkin~Nicholas J.
\newblock 2008.
\newblock {Query reformulation, search performance, and term suggestion devices
  in question-answering tasks}.
\newblock In {\em Proceedings of the IIiX '08}, pages 21--26, New York, NY.
  ACM.

\bibitem[\protect\citename{Yu and Qiang}2017]{survey_MTL_2017}
Zhang Yu and Yang Qiang.
\newblock 2017.
\newblock A survey on multi-task learning.
\newblock {\em arXiv preprint arXiv:1707.08114}.

\bibitem[\protect\citename{Yuelin and J}2008]{Li2008a}
Li~Yuelin and Belkin~Nicholas J.
\newblock 2008.
\newblock A faceted approach to conceptualizing tasks in information seeking.
\newblock {\em Inf. Process. Manage.}, 44(6):1822--1837.

\end{thebibliography}

\end{document}